\documentclass[journal]{IEEEtran}

\usepackage[T1]{fontenc}
\usepackage{amsmath,amssymb,bm}
\usepackage{booktabs}
\usepackage{graphicx}
\usepackage{cite}
\usepackage{xcolor}
\usepackage{url}
\usepackage{array}
\usepackage{tabularx}
\usepackage{placeins}
\usepackage{needspace}
\usepackage{stfloats}
\usepackage{microtype}
\usepackage[hidelinks]{hyperref}

\newcommand{\vect}[1]{\bm{#1}}
\newcommand{\pos}[1]{\left\langle #1\right\rangle_{+}}
\newif\ifshowrevisions
\showrevisionsfalse
\ifshowrevisions
  \colorlet{revisioncolor}{red}
\else
  \colorlet{revisioncolor}{black}
\fi
\DeclareRobustCommand{\rev}[1]{\textcolor{revisioncolor}{#1}}
\newenvironment{revblock}{\par\begingroup\color{revisioncolor}}{\par\endgroup}
\newif\ifshowreviewchanges
\showreviewchangesfalse
\ifshowreviewchanges
  \colorlet{reviewchangecolor}{red}
\else
  \colorlet{reviewchangecolor}{black}
\fi
\DeclareRobustCommand{\reviewchange}[1]{\textcolor{reviewchangecolor}{#1}}
\newcolumntype{Y}{>{\raggedright\arraybackslash}X}
\graphicspath{{figures/}}
\hypersetup{
 pdftitle={STC-MPM: Coupled Deformation, Progressive Damage, and Cut Formation in Soft-Tissue Cutting},
 pdfauthor={Kewei Zuo, Jiahe Chen, Ziao Kuang, Junsheng Zhu, and Etsuko Kobayashi}}

\title{STC-MPM: Coupled Deformation, \reviewchange{Progressive} Damage, and \reviewchange{Cut Formation} in Soft-Tissue Cutting}
\author{Kewei Zuo, Jiahe Chen, Ziao Kuang, Junsheng Zhu, and Etsuko Kobayashi%
\thanks{The authors are with the Department of Precision Engineering,
The University of Tokyo, Tokyo, Japan.}}

\begin{document}
\maketitle
\markboth{STC-MPM: Coupled Deformation, \reviewchange{Progressive} Damage, and \reviewchange{Cut Formation}}{Zuo et al.}
\raggedbottom

\begin{abstract}
\rev{Cutting is a recurring operation in robotic automation, from food
preparation to surgical tissue resection. For highly compliant targets,
blade motion may deform or displace the material rather than advance
the intended cut, while progressive failure changes load transfer and
subsequent tool--tissue interaction. A computational description must
therefore connect cut formation to the evolving mechanical response,
rather than specifying the incision independently of material failure.
\reviewchange{We present STC-MPM (Soft Tissue Cutting with the Material Point Method),
a framework that couples finite-strain deformation, history-driven
continuum damage, and configurable post-failure treatment. Damage
initiates only when a tensile-strain history
exceeds a material threshold within the blade process zone.
Progressive degradation changes stress transmission before the
post-failure policy is applied, while the remaining tissue continues
to deform, move, and interact with the tool. Numerical scalpel studies
using delayed particle deactivation follow this response through
insertion and withdrawal. Under identical
blade motion, reducing the damage-rate limit leaves the first recorded
damage time unchanged but delays and increases peak reaction force,
delays particle deactivation, and increases the fixed-cohort
displacement statistic at maximum insertion. The matched comparison
illustrates how post-initiation failure evolution alters subsequent
tool loading and recorded material motion. STC-MPM thus supports joint
analysis of cut formation and accompanying tissue response without a
pre-inserted cutting interface.}}
\end{abstract}


\section{Introduction}
\label{sec:introduction}

\IEEEPARstart{C}{utting} is a recurring operation in robotic automation,
from slicing fruits and vegetables in food preparation to surgical
tissue resection~\cite{jamdagni2024cutting,mu2024dexterous,ge2024astr}.
These tasks require a robot to produce an intended material separation,
not merely execute a blade trajectory. When a workpiece is well
supported and its shape changes little, tool travel can approximate
cutting progress. That approximation becomes unreliable for highly
compliant or weakly constrained targets: the tool can compress, stretch,
or displace the material before separation occurs. In surgery, the
intended cut must also be achieved while preserving the surrounding
tissue~\cite{ge2024astr}.

The central difficulty is the two-way coupling between cutting and
material response. Tool contact deforms the tissue and moves it
relative to the blade; progressive failure then reduces load-carrying
capacity and redistributes the loading. These changes alter subsequent
contact, deformation, and tissue motion. Blade advancement therefore
need not imply equivalent cut advancement, and the same tool trajectory
can produce different cutting and tissue-motion outcomes. Force alone
does not resolve this ambiguity. In automatic tissue traction, for
example, insufficient traction can impede cutting, whereas excessive
traction can cause unwanted splitting~\cite{liu2024traction}. A lower
force is therefore not an unconditional improvement, and smaller tissue
motion does not necessarily imply that an equivalent cut has been
achieved.

Addressing this difficulty computationally requires deformation, load
transfer, progressive loss of load-carrying capacity, and \reviewchange{cut formation
to participate in the same evolving interaction.} Material properties,
support conditions, and instrument motion provide the context for
interpreting those responses. Cutting models have supported the
analysis of tool loads and motion in deformable
materials~\cite{jamdagni2024cutting}, while differentiable simulation
has enabled parameter inference and motion
optimization~\cite{heiden2021disect,heiden2023disect}. Here, the aim is to
compare cut formation, local deformation, and region displacement
under controlled changes in the formulation or operating condition.
This requires obtaining tool loading and tissue response from the
same evolving body, rather than interpreting a force trace or final
incision in isolation.

Existing approaches make different assumptions about how a cut is
introduced. Geometric cutting algorithms can accommodate complex
cutting surfaces, but a separation rule based on the tool geometry does
not itself determine whether the material has reached a failure
condition~\cite{koschier2017cutting,ou2025cressim}. Models constructed
around a prepared cutting interface provide a different route, but
restrict separation to the surface represented in that
construction~\cite{heiden2023disect}. Mechanical failure models already
exist for tissue insertion and cutting~\cite{oldfield2013cohesive,jamdagni2024cutting};
the issue is therefore not an absence of failure mechanics across the
field. Our focus is a formulation that carries the failure history with
the deforming material, \reviewchange{drives post-failure treatment} through that response, and
continues to compute the resulting tissue motion without a pre-inserted
cutting interface.

We present STC-MPM (Soft Tissue Cutting with the Material Point Method)
to address this coupled process. Damage is initiated within the blade
process zone only when the material strain history exceeds its
threshold. \reviewchange{Progressive degradation changes stress transmission before
a distinct post-failure policy changes material activity or
particle--grid transfer compatibility.} These changes enter the
subsequent mechanical update, allowing failure to
alter later tool loading and tissue movement. Cut formation, tool
reaction, and tissue movement are thus computed together, rather than
prescribing the tissue trajectory or the final cut independently.
The material point method provides the particle--grid representation
used to advance the body and retain its material
history~\cite{sulsky1994particle}.

The main contribution is the STC-MPM formulation and its coupled
numerical realization for soft-tissue cutting. \rev{The numerical study
combines local damage-update checks with a representative scalpel
sequence, including withdrawal, and a matched failure-evolution
comparison. These studies examine tool loading, local failure, and
material-region motion together rather than treating a damage image or
final opening as a complete account of the interaction.}
\rev{Sections~\ref{sec:related} and~\ref{sec:preliminaries} review prior
work and background. Section~\ref{sec:framework} presents STC-MPM,
followed by its numerical assessment.}

\section{Related Work}
\label{sec:related}

\begin{revblock}
Two choices organize the comparison: the mechanical law governing
failure, and the computational representation of deformation and
separation. Neither a mesh nor a particle representation uniquely
determines the fracture law. Table~\ref{tab:method_comparison} compares
representative combinations of these choices.
\end{revblock}

\subsection{\rev{Mechanical Failure Models}}

\begin{revblock}
Cohesive-zone models (CZM) represent failure through a traction--separation
relation on an interface. They provide a direct description of crack
opening and interfacial work, as demonstrated in cohesive needle-insertion
models~\cite{oldfield2013cohesive}. A pre-inserted cohesive surface limits
failure to the represented path, but this restriction does not apply to
all CZM formulations: adaptively introduced cohesive surfaces can permit
fracture along evolving paths~\cite{camacho1996impact}. The relevant
tradeoff is how the fracture interface is introduced and updated, rather
than whether interface-based models include failure mechanics.

Continuum damage mechanics (CDM) instead represents degradation within
material volumes through internal state variables~\cite{lemaitre1985damage}.
For the present problem, its main advantage is the ability to describe
loss of load-carrying capacity around a moving blade while the affected
material remains part of the deforming body. This makes pre-separation
stress redistribution and its influence on surrounding tissue explicit,
without embedding the future cut as an interface in the initial body.
However, volumetric degradation does not by itself create independent
crack faces. Local softening also requires control of its spatial scale;
crack-band approaches relate the softening energy density to a fracture
energy through a characteristic length~\cite{bazant1983crackband}.

Phase-field fracture provides a gradient-regularized alternative closely
related to damage mechanics: an evolving field represents a diffuse
crack, with fracture energy and an internal length entering the governing
problem~\cite{miehe2010phasefield}. Finite-strain, anisotropic formulations
have been applied to soft biological tissue~\cite{gultekin2018anisotropic}.
These models offer capabilities beyond a local scalar damage law, but
also require spatial resolution of the regularized crack field. CDM,
CZM, and phase-field formulations therefore make different commitments
about degradation, localization, and crack representation; none is
universally preferable.

STC-MPM uses local, history-dependent volumetric damage to follow
progressive changes in tissue support, with \reviewchange{post-failure topology
handled separately.} This is a modeling choice for the coupled
cutting process, not a claim of a new general fracture theory. More
broadly, robotic cutting models already connect fracture toughness,
deformation, blade geometry, and motion~\cite{jamdagni2024cutting,mu2024dexterous}.
The distinction sought here concerns where failure is permitted, what
state drives its progression, and how that progression changes the
continuing mechanical response.
\end{revblock}

\begin{table*}[!t]
\centering
\caption{\rev{Methodological comparison of representative fracture and cutting formulations.}}
\label{tab:method_comparison}
\begingroup
\color{revisioncolor}
\footnotesize
\renewcommand{\arraystretch}{1.13}
\setlength{\tabcolsep}{3.2pt}
\begin{tabularx}{\textwidth}{@{}>{\raggedright\arraybackslash}p{1.80cm} >{\raggedright\arraybackslash}p{2.40cm} >{\raggedright\arraybackslash}p{2.20cm} >{\raggedright\arraybackslash}p{2.65cm} >{\raggedright\arraybackslash}p{3.05cm} Y@{}}
\toprule
Method & Failure initiation & Failure localization & Post-initiation evolution & Separation representation & Tool--failure coupling \\
\midrule
Ionescu et al.\ \cite{ionescu2006failure} & Matrix shear-strain and fiber-strain thresholds & Bulk material satisfying component thresholds & Abrupt loss of matrix/volumetric or fiber stress support & Component-level loss of support; no dedicated sharp cut partition & Penetration loads induce material failure; no cutting-edge initiation gate \\[2pt]
Cohesive FEM examples \cite{oldfield2013cohesive,moreno2026cutting} & Interface traction--separation law & Represented cohesive surface & Progressive loss of cohesive traction & Opening across the cohesive interface & Tool contact loads the deforming body and represented fracture surface \\[2pt]
Cutting XFEM \cite{koschier2017cutting} & Geometric cut specification, not a constitutive onset law & Tool-defined surface within elements & Discontinuity update; no damage law implied by enrichment & Enriched displacement field without conforming remeshing & Cut geometry determines the represented discontinuity \\[2pt]
MLS-MPM/CPIC cutting \cite{hu2018mlsmpm} & Geometric compatibility & Prescribed moving surface & Compatibility updates, rather than material degradation & Side-compatible particle--grid transfers & Tool/surface geometry defines the transfer discontinuity \\[2pt]
PFF-MPM branch of CD-MPM \cite{wolper2019cdmpm} & Tensile-energy-driven variational fracture & Evolving bulk phase field & Gradient-regularized damage with fracture energy and length scale & Diffuse stiffness degradation; not a cutter-side partition & General externally loaded fracture; no cutter-specific initiation gate \\[2pt]
DiSECt \cite{heiden2021disect,heiden2023disect} & Knife-contact-force-driven weakening of cutting springs & Preprocessed cutting surface & Interface spring stiffness decreases toward zero & Virtual-node interface opens as springs lose support & Tool contact progressively weakens a prepared interface \\[2pt]
CRESSim-MPM \cite{ou2025cressim} & Cutter-defined side compatibility & Tool distance-field surface & Geometric transfer restriction; no constitutive damage evolution in this cut rule & Incompatible P2G/G2P transfers are excluded & Cutter geometry sets separation; other tool regions supply contact \\[2pt]
CulinaryCut-VLAP \cite{koh2026culinarycut} & Contact, approach-speed, and downward-motion gates & Blade band & Particle damage reduces elastic moduli & Stiffness reduction; connectivity/color postprocessing & Contact/motion-gated damage rather than a tissue strain-history threshold \\[2pt]
BladeMaster \cite{yang2026blademaster} & Online blade-side classification & Swept blade surface & Persistent side-label updates & Persistent transfer discontinuity; cut faces can re-contact & Geometry generates cut history, retained after blade withdrawal \\[2pt]
\reviewchange{STC-MPM} & Maximum tensile principal Green--Lagrange strain history & New initiation in blade process zone; history follows material & Rate-limited target damage and quadratic stress degradation & \reviewchange{Delayed deactivation; optional persistent side-compatible transfers and retained failed phase} & Blade zone permits initiation; tissue state determines whether and when it occurs \\
\bottomrule
\end{tabularx}
\par\vspace{2pt}
\begin{minipage}{\textwidth}\scriptsize
The rows describe the cited constructions, not all possible extensions of each method.
PFF-MPM is one branch of CD-MPM. Geometric side classification does not by itself specify a
constitutive failure law. \reviewchange{The STC-MPM row uses the principal-strain setting. Its optional
transfer treatment is described in Section~\ref{sec:separation}; the archived scalpel
results use delayed deactivation without that barrier.}
\end{minipage}
\endgroup
\end{table*}

\Needspace{7\baselineskip}
\subsection{\rev{Computational Models for Soft-Material Cutting}}

\subsubsection{\rev{Mesh-Based Methods}}
\begin{revblock}
Finite-element methods (FEM) provide a well-established setting for
nonlinear tissue deformation and instrument interaction. Cohesive FEM
has reproduced needle insertion into soft phantoms~\cite{oldfield2013cohesive},
and inverse FEM has supported flexible-needle control in deforming
structures~\cite{adagolodjo2019needle}. Extended FEM can represent complex
cutting surfaces without conforming remeshing by enriching the
displacement field~\cite{koschier2017cutting}. Phase-field FEM offers a
different route, resolving diffuse cracks without explicitly tracking a
sharp interface~\cite{miehe2010phasefield,gultekin2018anisotropic}.

DiSECt combines finite-element deformation with virtual nodes and
springs across a preprocessed cutting surface. Knife contact forces
progressively weaken these springs, enabling differentiable parameter
inference and cutting-motion optimization~\cite{heiden2021disect,heiden2023disect}.
Its prepared fracture surface is a restriction of that particular
construction, not of FEM in general. A recent study by Moreno-Mateos and
Steinmann couples finite-strain bulk deformation with cohesive fracture,
adhesion, and friction, and compares the resulting soft-material cutting
response with experiments~\cite{moreno2026cutting}. Thus, coupled
mechanical cutting is not exclusive to particle-based methods.

The difficulty for conventional continuous, mesh-conforming FEM is that
changing material connectivity and large distortion require additional
handling of the discretization. Remeshing, enrichment, cohesive-interface
updates, or a resolved diffuse crack field address different parts of
this difficulty. Recent adaptive phase-field work also targets the cost
of resolving evolving fracture zones in soft, nearly incompressible
materials~\cite{dahal2026coarsening}. The motivation for a particle--grid
representation is therefore to avoid advecting a permanently connected
material mesh through the cutting process, not an inability of FEM to
simulate fracture.
\end{revblock}

\subsubsection{\rev{Particle-Based and Particle--Grid Methods}}
\begin{revblock}
Meshfree representations have long supported fracture without a fixed
volumetric mesh; Pauly et al.\ combined point-based deformation with
explicit crack tracking~\cite{pauly2005meshless}. MPM instead transports
material state on particles while solving momentum on a background
grid~\cite{sulsky1994particle}; APIC improves the velocity
transfer~\cite{jiang2015apic}. These are computational ingredients, not
constitutive failure criteria.

Mechanical failure also predates recent MPM cutting simulators. Ionescu
et al.\ modeled anisotropic soft-tissue failure by removing matrix or
fiber stress contributions after strain thresholds were
exceeded~\cite{ionescu2006failure}. The phase-field branch of CD-MPM
couples continuum damage and dynamics~\cite{wolper2019cdmpm}, while
AnisoMPM extends damage modeling to anisotropic
fracture~\cite{wolper2020aniso}. Other phase-field MPM formulations address
anisotropic surface energy~\cite{kakouris2019phasefield} and nearly
incompressible rubber-like solids~\cite{zhang2023mixed}. MPM has also been
used specifically for needle--tissue interaction~\cite{li2021needle}.
These studies preclude a novelty claim based simply on combining MPM,
soft materials, and mechanical failure.

A separate line of work concerns numerical discontinuities. Compatible
particle-in-cell (CPIC) transfers permit geometry-defined cuts and rigid
coupling~\cite{hu2018mlsmpm}; CRESSim-MPM uses tool distance fields to
restrict incompatible transfers in surgical cutting and
suturing~\cite{ou2025cressim}. BladeMaster generates persistent particle-side
labels online in total-Lagrangian MPM, maintaining discontinuities after
withdrawal and supporting subsequent contact between cut
faces~\cite{yang2026blademaster}. These developments address how material
regions remain kinematically distinct, which is different from specifying
a constitutive law for pre-separation degradation.

Not all recent MPM cutting models are geometry-only. CulinaryCut-VLAP
updates particle damage in a blade band using contact, approach-speed,
and downward-motion conditions, and reduces elastic moduli with
that damage~\cite{koh2026culinarycut}. Its driver differs from the tissue
strain-history threshold used here, but its damage-dependent stiffness
must be acknowledged. More generally, particle methods do not remove
all discretization difficulties: benchmarks identify challenges near
free surfaces and strong gradients~\cite{wolper2024surfaces}.

STC-MPM focuses on the coupling of blade-localized mechanical
initiation, history-driven stress degradation, and the subsequent
loading and motion of the remaining tissue. Unlike pre-inserted-interface
methods, its damage state resides in the moving material; unlike a
geometry-only cut rule, edge proximity alone is insufficient to cause
failure. \reviewchange{Post-failure handling includes delayed particle deactivation
and an optional persistent side-compatible transfer treatment
(Section~\ref{sec:separation}). The archived scalpel comparison
evaluates the delayed-deactivation configuration.}
\end{revblock}

\section{Preliminaries}
\label{sec:preliminaries}

\begin{figure*}[!t]
\centering
\includegraphics[width=\textwidth]{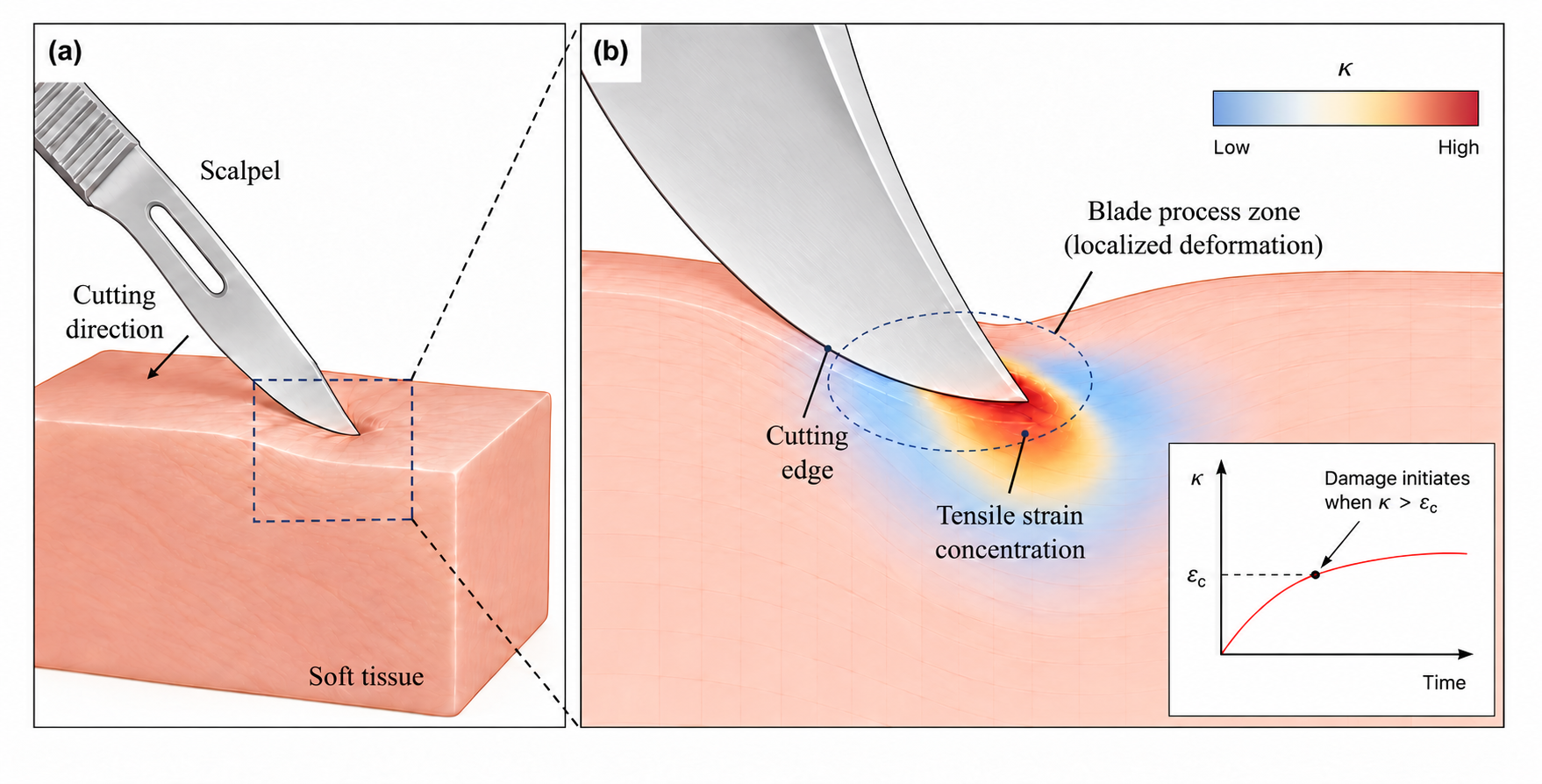}
\caption{Localized mechanics at cutting onset. (a) A scalpel loads a deformable
 tissue block. (b) The enlarged edge neighborhood illustrates concentration of
 the tensile-strain history $\kappa$, with the inset showing the initiation
 threshold $\varepsilon_c$. Tissue is still continuous at damage onset.
 The colors and inset are qualitative, not computed fields or measured data;
 new damage also requires the blade-process-zone condition in
 Section~\ref{sec:damage}.}
\label{fig:concentration}
\end{figure*}

\subsection{Tissue Deformation and Motion}

Let $\vect{X}$ and $\vect{x}=\boldsymbol{\chi}(\vect{X},t)$ denote the
reference and current positions of a material point. Its displacement,
deformation gradient, and volume ratio are
\begin{equation}
 \vect{u}=\vect{x}-\vect{X},\qquad
 \vect{F}=\frac{\partial\boldsymbol{\chi}}{\partial\vect{X}},\qquad
 J=\det\vect{F}>0.
 \label{eq:kinematics}
\end{equation}
Displacement and strain describe different aspects of tissue response.
A uniform translation changes position without changing $\vect{F}$.
Region motion is therefore reported alongside, rather than substituted
for, local deformation.

The intact tissue is represented by an isotropic compressible
Neo--Hookean solid. Its elastic energy per reference volume is
\begin{equation}
 \psi_0(\vect{F})=
 \frac{\mu}{2}\!\left[J^{-2/3}\operatorname{tr}
 (\vect{F}\vect{F}^{\mathsf T})-3\right]
 +\frac{K}{2}(\ln J)^2,
 \label{eq:elasticity}
\end{equation}
where $\mu=E_Y/[2(1+\nu)]$ and $K=E_Y/[3(1-2\nu)]$.
Here $E_Y$ and $\nu$ are Young's modulus and Poisson's ratio. The intact
Kirchhoff stress is obtained as
$\vect{\tau}_0=(\partial\psi_0/\partial\vect{F})\vect{F}^{\mathsf T}$.
This elastic baseline supplies the stress that is subsequently modified
by damage.

\subsection{\rev{Localized Tool Loading}}

A cutting tool produces nonuniform loading and deformation near its
edge~\cite{jamdagni2024cutting}. Figure~\ref{fig:concentration} illustrates
this interaction using a scalpel and an enlarged edge neighborhood.
The qualitative field and threshold inset distinguish local strain
concentration from the onset of damage: the tissue can remain continuous
while its tensile-strain history reaches the initiation threshold.
Stress and strain need not have coincident maxima, and the most highly
compressed contact region is not necessarily the region of greatest
tensile strain.

\subsection{\rev{Continuum Damage and \reviewchange{Post-Failure Treatment}}}
\label{sec:cdm_background}

\begin{revblock}
In a scalar continuum-damage description, $d\in[0,1]$ is an internal
material variable representing loss of load-carrying capacity; it is not
a crack opening or a tissue displacement~\cite{lemaitre1985damage}.
Irreversibility means that the accumulated damage cannot decrease during
subsequent unloading. A constitutive degradation rule maps this history
to a modified stress response. The particular driver and stress split
used in STC-MPM are specified in Section~\ref{sec:damage}.

\reviewchange{Damage initiation, attainment of a failure threshold, and
post-failure treatment are distinct. The damage variable controls
constitutive degradation but does not by itself define a sharp
displacement discontinuity. In STC-MPM, a post-failure policy determines
whether failed particles remain active or become inactive, while an
optional transfer-compatibility rule restricts particle--grid coupling
across a resolved cut.}

\reviewchange{For local softening, a crack-band construction relates fracture energy
per unit area to a softening energy density through a characteristic
length~\cite{bazant1983crackband}. STC-MPM uses this scaling as a reference
for the damage-evolution interval, with a reference fracture-energy
parameter $G_c^\ast$ and characteristic length $l_c$. The nominal scale
$G_c^\ast/l_c$ is distinguished from the energy dissipated along the
simulated loading path. Fracture energy, tool work, and energy removed
with deactivated material must therefore be distinguished~\cite{taylor2012toughness}.}
\end{revblock}

\subsection{Material Point Representation}

MPM stores motion and material history on particles while advancing
momentum on a background grid~\cite{sulsky1994particle}. The grid is
reset between steps, whereas the particle history is retained.
\reviewchange{STC-MPM uses quadratic B-spline interpolation to provide
smooth particle--grid weights. The affine particle-in-cell (APIC)
transfer retains local affine velocity information, reducing the
excessive numerical dissipation of conventional particle-in-cell (PIC)
transfer~\cite{jiang2015apic}.} Interpolation is evaluated in the current
configuration. These components provide the discretization, not a
failure criterion.

Particle separation must also be distinguished from particle--grid
coupling. Nearby particles can share a grid state even when a cut is
visible between them. Transfer-compatible formulations address this
issue by restricting which contributions can share or recover a nodal
velocity~\cite{hu2018mlsmpm}. This distinction is revisited with the
current post-failure treatment in Section~\ref{sec:separation}.

\FloatBarrier
\section{STC-MPM Framework}
\label{sec:framework}

\subsection{Coupled Formulation}

\begin{figure*}[!t]
\centering
\includegraphics[width=\textwidth]{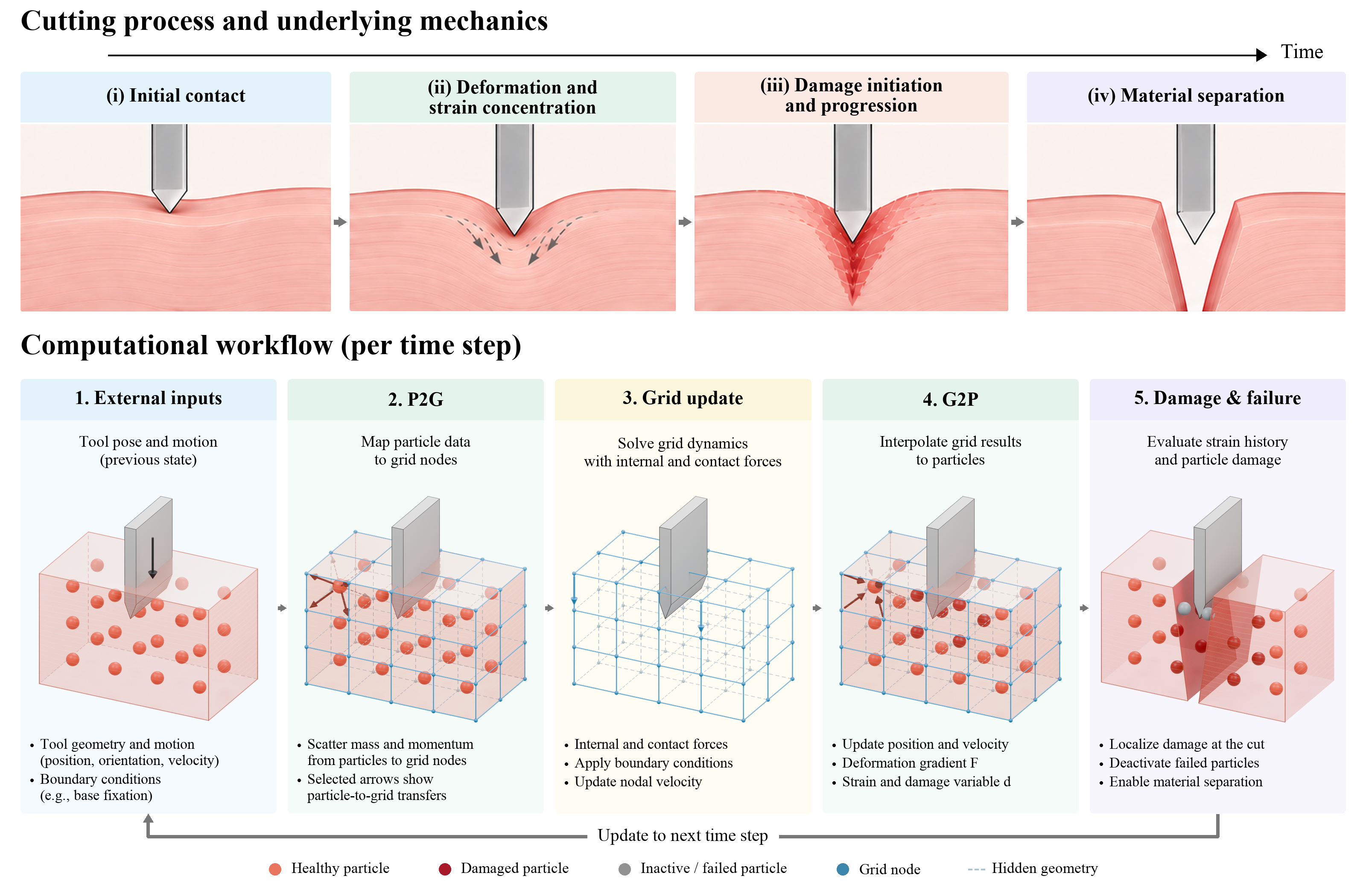}
\caption{Cutting process and STC-MPM update cycle. Top: initial contact,
 deformation and strain concentration, damage initiation and progression,
 and \reviewchange{cut formation by failed-particle deactivation.} Bottom: external inputs, particle-to-grid (P2G)
 transfer, grid dynamics and contact, grid-to-particle (G2P) transfer, and
 damage/failure updates within one time step. Damage is evaluated using
 the updated deformation state; degraded stress and active-set changes
 affect the subsequent step. The numerical cycle is repeated within each
 physical stage, not mapped one-to-one onto the upper panels. All views
 are schematic.}
\label{fig:overview}
\end{figure*}

The inputs are tissue and cutter geometries, material and contact
properties, support conditions, and prescribed cutter motion. The tissue
trajectory and final opening are not prescribed independently.
Figure~\ref{fig:overview} follows the same material neighborhood through
the coupled calculation. Initially, the tissue carries the tool load
while deforming and moving. After damage starts, the affected material
remains active with a progressively degraded response. This changes the
stress driving subsequent motion. \reviewchange{A distinct post-failure policy
changes material activity or transfer compatibility, while the
remaining material continues to interact with the cutter.}

\reviewchange{The distinction between constitutive damage and post-failure topology
is central: a damage update is not an immediate geometric cut.} Tool
reaction and region motion are recorded throughout this process.
They are computed consequences of the evolving body, not an extra
stage that begins only after \reviewchange{a post-failure topology update.}

\subsection{Tool--Tissue Interaction}

The rigid cutter \rev{can translate and rotate}. Its surface velocity at a
contact point $\vect{x}$ is
\begin{equation}
 \vect{u}_c(\vect{x},t)=\vect{v}_c(t)
 +\vect{\omega}_c(t)\times(\vect{x}-\vect{x}_c(t)),
 \label{eq:tool_velocity}
\end{equation}
where $\vect{x}_c$, $\vect{v}_c$, and $\vect{\omega}_c$ are its reference
position, translational velocity, and angular velocity. A signed-distance
representation supplies the contact surface and normal. Non-penetration
and Coulomb friction are imposed using relative tool--tissue velocity.
\reviewchange{The contact impulse is the contact-induced change in
tissue momentum. Its tangential component $\Delta\vect{p}_{t,a}$
represents friction in the local tangent plane. Coulomb friction imposes
$\|\Delta\vect{p}_{t,a}\|\le\mu_f\Delta p_{n,a}$, where $a$ indexes the
applied contact impulses, $\Delta p_{n,a}\ge0$ is the compressive
normal-impulse magnitude, and $\mu_f$ is the dimensionless friction
coefficient.}

All contacting tool surfaces can load or drag tissue; this region is
distinct from the smaller region in which new cutting damage is
permitted. With $\Delta\vect{p}_a$ the contact impulse applied to the
tissue at $\vect{x}_a$, the reaction on the tool is
\begin{equation}
 \begin{aligned}
 \vect{F}_c&=-\frac{1}{\Delta t}\sum_a\Delta\vect{p}_a,\\
 \vect{M}_c&=-\frac{1}{\Delta t}\sum_a
 (\vect{x}_a-\vect{x}_c)\times\Delta\vect{p}_a.
 \end{aligned}
 \label{eq:reaction}
\end{equation}
\reviewchange{Here $\Delta t$ is the simulation time step, and
$\vect{F}_c$ and $\vect{M}_c$ are the step-averaged reaction force and
moment on the cutter, with the moment taken about $\vect{x}_c$.
The negative signs give the reaction opposite to the impulse applied
to the tissue; friction is already included in $\Delta\vect{p}_a$.}
Each applied impulse is counted once. These are reaction outputs, not a
separately imposed cutting-force law. The cutter motion remains
prescribed in this study.

\subsection{Mechanically Triggered Progressive Failure}
\label{sec:damage}

For particle $p$, the tensile strain measure is the positive maximum
principal Green--Lagrange strain,
\begin{equation}
 \vect{E}_{G,p}=\tfrac12(\vect{F}_p^{\mathsf T}\vect{F}_p-\vect{I}),
 \qquad
 \varepsilon_{1,p}=\pos{\lambda_{\max}(\vect{E}_{G,p})},
 \label{eq:principal}
\end{equation}
where $\pos{a}=\max(a,0)$. Its history is irreversible:
\begin{equation}
 \kappa_p^{n+1}=\max(\kappa_p^n,\varepsilon_{1,p}^{n+1}).
 \label{eq:history}
\end{equation}
New damage requires both proximity to the cutting edge and a history
exceeding the critical strain $\varepsilon_c$. Proximity alone does not
cause failure. After initiation, the history follows the material and
can continue to evolve after the edge moves away. No cutting interface
is inserted into the initial tissue configuration.

Damage $d_p\in[0,1]$ reduces the stress response through
\begin{equation}
 g(d)=r+(1-r)(1-d)^2,
 \label{eq:degradation}
\end{equation}
\reviewchange{where $r$ is a prescribed dimensionless residual-stiffness
fraction. Since $g(0)=1$ and $g(1)=r$, it keeps the multiplier of the
degraded stress components nonzero; volumetric compression is retained
separately. The reference value $r=0.05$ is a numerical setting, not a
tissue-calibrated parameter. Compression-only projection and particle
deactivation are separate post-failure operations, as described in
Section~\ref{sec:separation}.}
\begin{revblock}
The implementation used in this study evolves damage toward a target
that increases linearly with strain history:
\begin{equation}
 \begin{aligned}
 \varepsilon_f&=\varepsilon_c+\reviewchange{\frac{2G_c^\ast}{l_c E_Y\varepsilon_c}},\\
 d_p^*&=\operatorname{clamp}\!\left(
 \frac{\kappa_p^{n+1}-\varepsilon_c}{\varepsilon_f-\varepsilon_c},0,1\right).
 \end{aligned}
 \label{eq:damage_target}
\end{equation}
\reviewchange{Here $G_c^\ast$ is the reference fracture-energy parameter, with units
of $\mathrm{J/m^2}$, and sets the target-damage interval through the
characteristic length $l_c$; the reference case uses $l_c=\Delta x$.}
After initiation, the update enforces irreversibility and a
physical-time rate limit,
\begin{equation}
 d_p^{n+1}=\min\!\left\{\max(d_p^n,d_p^*),\,
 d_p^n+\dot d_{\max}\Delta t\right\},
 \label{eq:damage_rate}
\end{equation}
\reviewchange{where $\dot d_{\max}$ is the damage-rate limit in
$\mathrm{s^{-1}}$. It caps the damage increment at
$\dot d_{\max}\Delta t$ per step, controlling how rapidly constitutive
degradation can accumulate when the target permits further growth.
This model setting is distinct from strain rate and from the
post-threshold erosion delay, and is not an experimentally calibrated
tissue-fracture kinetic parameter.}
This linear target-damage relation, combined with
Eq.~\eqref{eq:degradation}, is not a linear post-peak stress law.
\end{revblock}

Deviatoric stress and volumetric tension are multiplied by $g(d)$;
volumetric compression is retained. The resulting Kirchhoff stress
$\vect{\tau}_{d,p}$ enters the subsequent momentum update. This links
loss of load-carrying capacity to the tool reaction and tissue motion.
\rev{\reviewchange{Equation~\eqref{eq:damage_target} sets a reference interval; it does
not by itself enforce dissipation of $G_c^\ast$ per unit crack area under
the implemented quadratic degradation, stress split, and rate limit.
Subsequent particle deactivation additionally removes the energy
remaining in the failed material. Thus $G_c^\ast$ is not identified here
with experimentally measured tissue fracture toughness. The nominal
energy record, constitutive dissipation, tool work, and deactivation
losses must be interpreted as distinct quantities.}}

\subsection{Separation and Continued Interaction}
\label{sec:separation}

\reviewchange{A particle remains active under the damaged constitutive law while
$d_p<d_{\mathrm{fail}}$. Reaching $d_p\ge d_{\mathrm{fail}}$ marks it
as failed, but does not itself make it inactive or require $d_p=1$.
The updated solver provides two alternative post-failure policies.}

\begin{figure*}[!t]
\centering
\includegraphics[width=\textwidth]{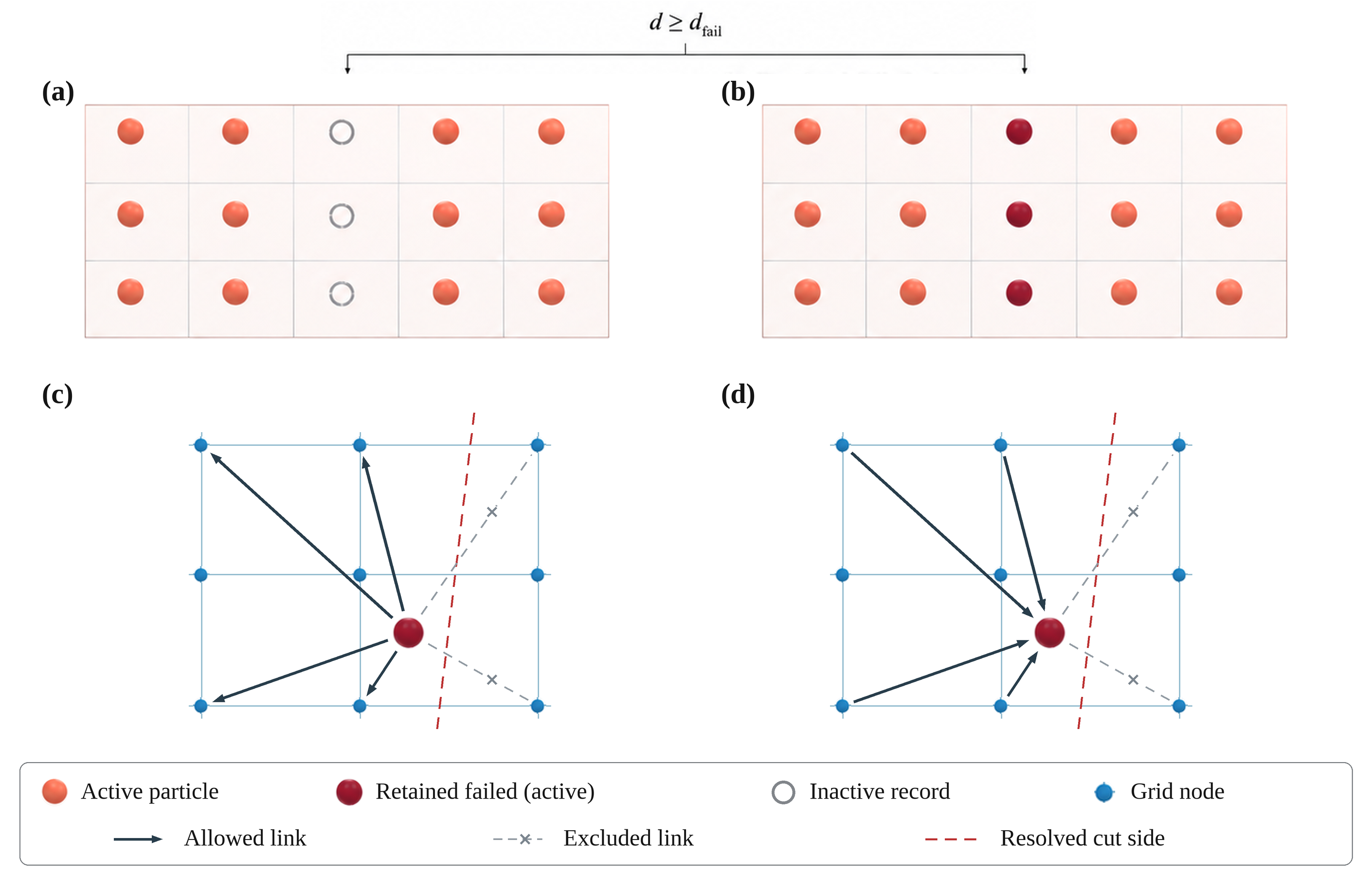}
\caption{\reviewchange{Post-failure policies and particle--grid transfer compatibility.
(a) Inactive records after the delayed-erosion interval.
(b) Retained failed particles that remain active with compression-only
recoverable elasticity. (c), (d) Representative P2G and G2P links for
the same retained failed particle with the optional transfer barrier
enabled. Dark arrows denote allowed transfers; crossed grey links are
excluded. The red dashed line indicates the resolved local cut side.
Panels (a), (b) compare alternative policies, whereas (c), (d) show
opposite transfer directions, not separate policies. Inactive records
do not participate in either transfer.}}
\label{fig:transfer}
\end{figure*}

\reviewchange{With \emph{delayed erosion} [Fig.~\ref{fig:transfer}(a)], a failed
particle remains active during a prescribed physical-time delay
measured from its first failure-threshold crossing. During this delay
it retains the damaged constitutive response, without a compression-only
projection. Once the delay expires, it becomes \emph{inactive}: its
mass, momentum, internal-force, and contact contributions are removed,
and it no longer participates in P2G or G2P. Its last state is retained
only as a record, represented by a hollow marker in the figure.}

\reviewchange{With the \emph{retained failed phase} selected [Fig.~\ref{fig:transfer}(b)],
threshold-failed particles instead remain active, preserving mass,
momentum, motion, and contact. Their recoverable elastic state is
projected to a \emph{compression-only pressure response}, without
recoverable tensile or shear support. This elastic projection is
separate from the total deformation used to track material history;
compression-only does not mean mechanically inactive. This policy
requires a resolved material cut side and the persistent transfer
barrier, and is mutually exclusive with particle removal. The two
policies are alternatives, not successive stages or a switch determined
by the instantaneous sign of stress.}

\reviewchange{For cutter-localized failure with a resolved side, the optional
persistent barrier applies the same particle-side compatibility mask
in P2G and G2P after the failure threshold is reached.
Figures~\ref{fig:transfer}(c), (d) show the two transfer directions for
the same active retained failed particle: compatible links are used
and incompatible links are excluded. The dashed boundary indicates
an already resolved local cut side, not a pre-inserted interface.
The truncated stencil is renormalized about its retained centroid,
with a regularized moment tensor for affine reconstruction. The grid
geometry remains unchanged; its nodal fields are reassembled from
eligible particle contributions. This single-field barrier is
experimental and disabled by default.}

\subsection{Particle--Grid Update}
\label{sec:update}

\reviewchange{The following equations describe the transfers without the optional
barrier, as used in the archived scalpel runs. Barrier-enabled
transfers use the compatible stencil and affine reconstruction
described in Section~\ref{sec:separation}.}

\reviewchange{Let $\mathcal{A}^n$ be the active particle set, including failed
particles awaiting erosion but excluding inactive records. Membership
is determined by particle activity, not by the failure threshold alone.
With grid node $i$, interpolation weight $w_{ip}=N_i(\vect{x}_p^n)$,
mass $m_p$, and APIC affine velocity $\vect{C}_p^n$, mass and momentum
are transferred as}
\begin{equation}
 \begin{aligned}
 m_i&=\sum_{p\in\mathcal{A}^n} w_{ip}m_p,\\
 \vect{q}_i&=\sum_{p\in\mathcal{A}^n} w_{ip}m_p
 \left[\vect{v}_p^n+\vect{C}_p^n(\vect{x}_i-\vect{x}_p^n)\right].
 \end{aligned}
 \label{eq:p2g}
\end{equation}
These are the standard APIC transfer quantities~\cite{jiang2015apic}.
Using the spatial weight gradient $\nabla w_{ip}$ and reference particle
volume $V_p^0$, the internal force and unconstrained nodal update are
\begin{equation}
 \begin{aligned}
 \vect{f}_i^{\mathrm{int}}
 &=-\sum_{p\in\mathcal{A}^n}V_p^0\vect{\tau}_{d,p}^{n}\nabla w_{ip},\\
 \widetilde{\vect{v}}_i
 &=\frac{\vect{q}_i+\Delta t(\vect{f}_i^{\mathrm{int}}
                         +\vect{f}_i^{\mathrm{ext}})}{m_i},\quad m_i>0.
 \end{aligned}
 \label{eq:grid}
\end{equation}
External forces include the specified body loading. Supports and cutter
contact correct this velocity to $\vect{v}_i^{n+1}$.

The corrected grid field returns the particle motion,
\begin{equation}
 \vect{v}_p^{n+1}=\sum_i w_{ip}\vect{v}_i^{n+1},\qquad
 \vect{x}_p^{n+1}=\vect{x}_p^n+\Delta t\,\vect{v}_p^{n+1},
 \label{eq:g2p}
\end{equation}
and its local gradient advances the deformation,
\begin{equation}
 \begin{aligned}
 \vect{L}_p^{n+1}&=\sum_i\vect{v}_i^{n+1}\otimes\nabla w_{ip},\\
 \vect{F}_p^{n+1}&=(\vect{I}+\Delta t\,\vect{L}_p^{n+1})\vect{F}_p^n.
 \end{aligned}
 \label{eq:F_update}
\end{equation}
The affine velocity is reconstructed by the APIC update. The resulting
strain updates material history and damage; the next mechanical step
uses the degraded stress and updated active set. The corrected motion field thus advances position, deformation, and
subsequent material history within the same update.

\subsection{Tool and Tissue Outputs}

\begin{revblock}
The framework provides tool force and moment, local strain and damage,
and material-region motion. In the present scalpel study, the tissue-side
motion statistic uses a fixed initial surface-particle cohort
$\mathcal{R}_0$ rather than a spherical target or a centroid trajectory.
Its median displacement magnitude is
\begin{equation}
 U_{50}(t)=\operatorname{median}_{p\in\mathcal{R}_0}
 \left\|\vect{x}^{\mathrm{rec}}_p(t)-\vect{X}_p\right\|,
 \label{eq:region_motion}
\end{equation}
where $\vect{x}^{\mathrm{rec}}_p(t)$ is the current recorded position of
an active particle and the last saved position of a deactivated particle.
The cohort membership remains fixed; deactivated particles are not
silently removed from this statistic. Consequently, $U_{50}$ is neither
a centroid trajectory of surviving tissue nor a strain measure, and its
post-withdrawal value alone does not quantify elastic recovery.

Tool work, constitutive energy loss, \reviewchange{nominal energy records,}
and energy removed with inactive material remain distinct quantities.
They are not treated as interchangeable measurements of fracture.
\end{revblock}

\section{Numerical Studies and Results}
\label{sec:numerical}

\subsection{Reference Configuration}
\label{sec:configuration}

\begin{revblock}
The reference configuration uses a scalpel blade acting on a
$4\times4\times4\,\mathrm{mm}$ tissue block with a fixed bottom layer
of $0.3\,\mathrm{mm}$. The reference run and the matched comparison use
the same v14.12 implementation, including the target-damage relation
and stress treatment in Section~\ref{sec:damage}. \reviewchange{These archived runs
use delayed erosion without the optional transfer barrier or retained
failed phase described in Section~\ref{sec:separation}.}

The blade starts with a $0.4\,\mathrm{mm}$ gap above the tissue.
The prescribed trajectory uses an insertion depth of $0.8\,\mathrm{mm}$,
an advance speed of $2\,\mathrm{mm/s}$, and simultaneous sliding at
$4\,\mathrm{mm/s}$. The complete motion, including withdrawal, lasts
$1.05\,\mathrm{s}$. The blade length, height, body thickness, and edge
thickness are $10$, $8$, $0.8$, and $0.2\,\mathrm{mm}$, respectively,
with a $3\,\mathrm{mm}$ bevel height.

Table~\ref{tab:parameters} gives the material and numerical settings.
For new damage, the blade-edge distance must not exceed
$0.3\,\mathrm{mm}$, and the cutter signed distance must not exceed
the run's contact-margin metadata plus the $0.25\,\mathrm{mm}$
activation distance. Initiated damage subsequently follows the
material history. \reviewchange{Each particle remains active during the
$0.5\,\mathrm{ms}$ interval following its first failure-threshold
crossing and is then deactivated. No compression-only retained phase
is used in these runs.}
\end{revblock}

\begin{table}[!htbp]
\centering
\caption{\rev{Reference scalpel settings; the main comparison changes only
$\dot d_{\max}$ from $10$ to $1\,\mathrm{s^{-1}}$.}}
\label{tab:parameters}
\renewcommand{\arraystretch}{1.04}
\begin{tabularx}{\columnwidth}{@{}Y l@{}}
\toprule
Quantity & Value\\
\midrule
Young's modulus, $E_Y$ & $60\,\mathrm{kPa}$\\
Poisson's ratio, $\nu$ & \rev{$0.45$}\\
\rev{Density, $\rho$} & \rev{$1000\,\mathrm{kg/m^3}$}\\
Critical strain, $\varepsilon_c$ & \rev{$0.22$}\\
\reviewchange{Reference fracture-energy parameter, $G_c^\ast$} & \rev{$0.3\,\mathrm{J/m^2}$}\\
\rev{Characteristic length, $l_c$} & \rev{$0.1\,\mathrm{mm}$}\\
Residual stiffness, $r$ & \rev{$0.05$}\\
Failure threshold, $d_{\mathrm{fail}}$ & \rev{$0.75$}\\
\rev{Damage-rate limit, $\dot d_{\max}$} & \rev{$10\,\mathrm{s^{-1}}$}\\
\rev{Post-threshold delay} & \rev{$0.5\,\mathrm{ms}$}\\
Friction coefficient, $\mu_f$ & \rev{$0.18$}\\
Grid spacing, $\Delta x$ & \rev{$0.1\,\mathrm{mm}$}\\
\rev{Particles per cell} & \rev{$27$}\\
\rev{Particle spacing} & \rev{$0.033333\,\mathrm{mm}$}\\
\rev{Initial particle count} & \rev{$1{,}728{,}000$}\\
Time step, $\Delta t$ & \rev{$\approx1.9\,\mu\mathrm{s}$}\\
\rev{Number of steps} & \rev{$552{,}632$}\\
\bottomrule
\end{tabularx}
\end{table}

\begin{figure*}[!t]
\centering
\includegraphics[width=\textwidth]{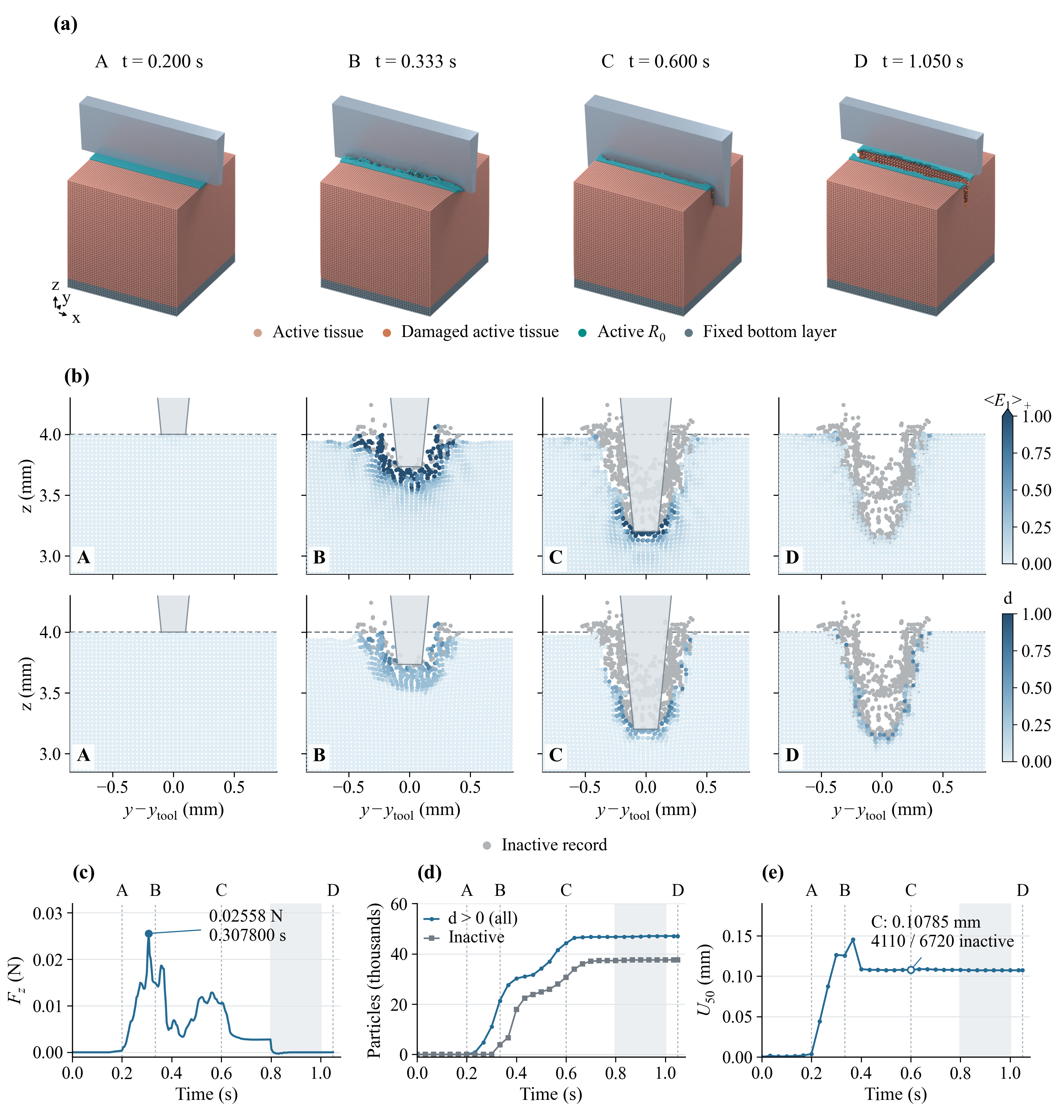}
\caption{\rev{Reference scalpel response with
$\dot d_{\max}=10\,\mathrm{s^{-1}}$. (a) Active-tissue configurations at
A--D: initial loading, damage evolution, maximum insertion, and after
withdrawal, at approximately $0.200$, $0.333$, $0.600$, and
$1.050\,\mathrm{s}$, respectively. (b) Local response at the same times:
positive maximum principal Green--Lagrange strain in the upper row and
damage $d$ in the lower row, in the fixed spatial section at
$x=2.0\,\mathrm{mm}$ with thickness $0.1\,\mathrm{mm}$. Grey crosses
indicate last recorded inactive positions. (c) Axial reaction $F_z$.
(d) Counts of particles with $d>0$ and inactive particles; these sets
are not disjoint. (e) Median displacement magnitude $U_{50}$ of the
fixed initial $6720$-particle surface cohort, including inactive records.
Letters on the histories identify the saved spatial frames, while the
curves cover the complete motion. Shading denotes prescribed withdrawal
from $0.8$ to $1.0\,\mathrm{s}$. The common display and sampling
conventions are given in Section~\ref{sec:configuration}.}}
\label{fig:reference}
\end{figure*}

\begin{figure*}[!t]
\centering
\includegraphics[width=\textwidth]{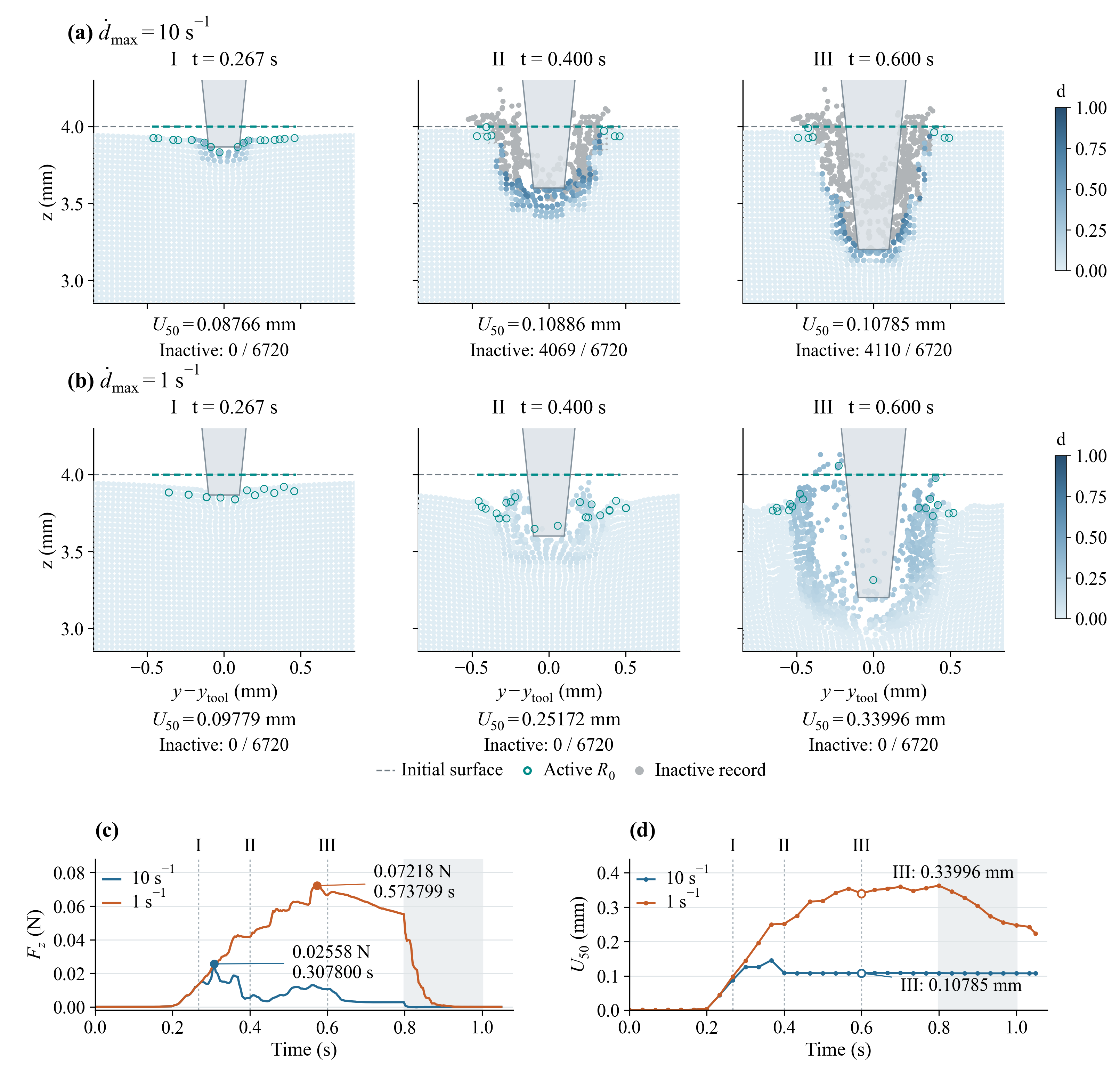}
\caption{\rev{Matched damage-rate comparison at $27$ particles per
cell under identical blade motion and otherwise matched settings.
(a), (b) Local damage and particle activity for
$\dot d_{\max}=10$ and $1\,\mathrm{s^{-1}}$, respectively. Columns
I--III show the same saved times, approximately $0.267$, $0.400$, and
$0.600\,\mathrm{s}$. Both rows use the fixed spatial section at
$x=2.0\,\mathrm{mm}$ with thickness $0.1\,\mathrm{mm}$ and the same
damage scale. Outlined markers identify displayed active members of the
initial surface cohort; grey crosses show last recorded inactive
positions. Values below each view report $U_{50}$ and the inactive
cohort count. (c) Axial reaction histories. (d) Median displacement
magnitude of the same fixed initial $6720$-particle cohort, retaining
inactive records. I--III on the histories identify the spatial samples;
shading denotes withdrawal from $0.8$ to $1.0\,\mathrm{s}$. The lower
rate limit delays peak resistance and increases the displacement
statistic at maximum insertion.}}
\label{fig:comparison}
\end{figure*}

\begin{revblock}
The reported tool-side component is axial reaction $F_z$, with positive
values denoting resistance to insertion. Tissue-side outputs include
strain, damage, particle activity, and $U_{50}$ in
Eq.~\eqref{eq:region_motion}. The fixed initial surface cohort
$\mathcal{R}_0$ contains $6720$ particles on the upper tissue surface
near the tool in each $27$-particle-per-cell run. Its membership remains
unchanged, including particles that subsequently become inactive;
$U_{50}$ uses their last recorded positions. No spherical target
tracking is used.

Each retained run has $33$ archived spatial frames and $278$ load-history
samples. Snapshot timestamps are rounded to three decimal places; the
underlying samples retain their saved times. Load histories and spatial
frames have different sampling schedules: the samples used near maximum
insertion are at approximately $0.600399\,\mathrm{s}$ and
$0.6000005\,\mathrm{s}$, respectively. Particle counts and $U_{50}$ are
computed at the spatial-frame times. Force-peak and first-event times
below refer to recorded samples rather than continuous-time event
localization. Shading in the histories identifies the prescribed
withdrawal interval, $0.8$--$1.0\,\mathrm{s}$.

For the display in Fig.~\ref{fig:reference}(a), active particles are rendered as opaque spherical glyphs with a fixed
$0.038\,\mathrm{mm}$ display radius. The glyph size is a visualization
setting, not a particle-volume or crack-width estimate; no crack surface
is fitted. All active cohort members are shown; deterministic display
sampling is applied outside the cohort. The blade geometry uses its
saved tip position and quaternion,
with one fixed world-space display crop and the same camera and lighting
across frames. The $4\,\mathrm{mm}$ cubic specimen and its fixed
$0.3\,\mathrm{mm}$ bottom layer are specified above.

The local views in Figs.~\ref{fig:reference}(b) and
\ref{fig:comparison}(a), (b) use a fixed spatial $yz$ section centered at
$x=2.0\,\mathrm{mm}$ with total thickness $0.1\,\mathrm{mm}$; particles
are selected by their recorded current positions rather than initial
material membership. Colored points show active-particle fields, and
grey crosses show last recorded inactive positions. Grey dashed lines
mark the initial tissue surface, and teal dashed segments mark the
initial cohort footprint; outlined cohort markers in
Fig.~\ref{fig:comparison} are a display sample. The strain field is the
positive part of the largest principal Green--Lagrange strain computed
from the total deformation gradient. Its color scale saturates at $1$
for display. Damage uses the same $0$--$1$ scale in all local views.
In Fig.~\ref{fig:reference}(d), the $d>0$ count includes inactive records,
so the damaged and inactive sets are not disjoint.
\end{revblock}

\Needspace{7\baselineskip}
\subsection{\rev{Basic Damage-Update Checks}}
\label{sec:verification}

\begin{revblock}
Local numerical checks assessed the principal-strain driver and the
physical-time damage-rate update. The analytic maximum-principal-strain
calculation agreed with a numerical eigensolver for the tested
deformation states, including repeated eigenvalues.

With a target damage of $0.9$ and a rate limit of
$2.5\,\mathrm{s^{-1}}$, time steps of $1$ and $0.2\,\mathrm{ms}$ both
produced a damage value of $0.5$ after $0.2\,\mathrm{s}$. These CPU
checks concern the local update; they are not a material-point
loading--unloading experiment, a full-simulation time-step convergence
study, or validation of the cutting response.
\end{revblock}

\subsection{Representative Cutting Case}
\label{sec:reference_results}

\begin{revblock}
The reference sequence includes insertion, sliding, and withdrawal.
Figure~\ref{fig:reference}(a) shows the global configurations at A--D,
while Fig.~\ref{fig:reference}(b) pairs each configuration with local
strain and damage. These frames link the reaction in
Fig.~\ref{fig:reference}(c), particle counts in
Fig.~\ref{fig:reference}(d), and fixed-cohort motion in
Fig.~\ref{fig:reference}(e). The observations distinguish damage onset
from deactivation and describe the continuing response after removal
begins.

Damage was first recorded at $0.2204\,\mathrm{s}$ and particle
deactivation at $0.2964\,\mathrm{s}$. The axial reaction reached a
recorded peak of $0.02558\,\mathrm{N}$ at $0.3078\,\mathrm{s}$, followed
by lower but continuing resistance during further insertion.
Near maximum insertion, $F_z$ remained approximately
$0.01044\,\mathrm{N}$, and the corresponding spatial frame gave
$U_{50}=0.10785\,\mathrm{mm}$ at frame C. At that frame, $4110$ of the
$6720$ cohort particles were already inactive, so more than half of the
displacement values entering the median use last recorded positions.
These records connect evolving failure with a changing load history and
recorded material displacement; the later $U_{50}$ plateau alone does
not establish residual deformation of the surviving tissue.

The interval between the first damage and first deactivation records
is not the configured removal delay, nor a common damage lifetime for
all particles. Likewise, a nonzero deactivation count does not establish
a fully severed body or a persistent open crack after withdrawal.
\end{revblock}

A change in reaction is interpreted with the corresponding configuration
and local response, not assigned to failure from force alone.
\rev{Strain and the fixed-cohort displacement statistic distinguish
local shape change from accumulated motion, although the latter also
contains last-saved inactive-particle positions.} Inactive volume is a
removal record, not a direct measure of completed separation.

\subsection{\rev{Effect of Failure Evolution}}
\label{sec:comparison}

\begin{revblock}
The controlled comparison changes only the damage-rate limit from
$10$ to $1\,\mathrm{s^{-1}}$ under identical blade motion and otherwise
matched settings. Both runs use STC-MPM; this is a model-sensitivity
comparison, not a competing-method benchmark.
Table~\ref{tab:rate_comparison} gives the retained numerical results.
Figure~\ref{fig:comparison}(a), (b) compares the local states at I--III,
with the baseline rate in the upper row and the lower rate in the lower
row. Figure~\ref{fig:comparison}(c), (d) gives the corresponding full
reaction and fixed-cohort displacement histories. The displayed spatial
states share times and views, whereas the force curves retain their
denser load-history sampling.

At $27$ particles per cell, the lower rate limit shifted the force peak
from $0.3078$ to $0.5738\,\mathrm{s}$ and increased its magnitude from
$0.02558$ to $0.07218\,\mathrm{N}$. At maximum insertion, $U_{50}$
increased from $0.10785$ to $0.33996\,\mathrm{mm}$ at frame III.
The baseline cohort contained $4110$ inactive particles at that time,
whereas the lower-rate cohort contained none. Thus the two values use
the same initial membership but different mixtures of current and
last-recorded positions. The first recorded damage occurred at
approximately $0.2204\,\mathrm{s}$ in both runs,
whereas first particle deactivation shifted from approximately
$0.2964$ to $0.9690\,\mathrm{s}$. Thus the observed difference concerns
post-initiation evolution, not a shift in damage onset.
\end{revblock}

\begin{table}[!htbp]
\centering
\caption{\rev{Matched damage-rate comparison. $F_{z,\mathrm{pk}}$ and
$t_{\mathrm{pk}}$ are the sampled force peak and its time; $U_{50}$ is
evaluated at the saved frame near maximum insertion.}}
\label{tab:rate_comparison}
\begingroup\color{revisioncolor}
\setlength{\tabcolsep}{3.8pt}
\renewcommand{\arraystretch}{1.08}
\begin{tabular}{@{}rrrrr@{}}
\toprule
PPC & $\dot d_{\max}$ & $F_{z,\mathrm{pk}}$ & $t_{\mathrm{pk}}$ & $U_{50}$\\
 & ($\mathrm{s^{-1}}$) & (N) & (s) & (mm)\\
\midrule
8  & 10 & 0.02353 & 0.3078 & 0.06712\\
8  & 1  & 0.05164 & 0.5206 & 0.30990\\
27 & 10 & 0.02558 & 0.3078 & 0.10785\\
27 & 1  & 0.07218 & 0.5738 & 0.33996\\
\bottomrule
\end{tabular}
\endgroup
\end{table}

\begin{revblock}
The $8$-particle-per-cell pair exhibits the same directional changes:
the peak moves to a later time, and both its magnitude and the local
displacement statistic increase. This repeats the trend at another
particle sampling; it does not establish particle-density or grid
convergence.

Changing the rate limit affects both stress degradation and the timing
of subsequent particle deactivation, so their individual contributions
are not isolated. The comparison shows that the evolution of failure
changes the duration of load support and the accumulated local motion
even when the blade trajectory is unchanged.
\end{revblock}

\rev{The matched comparison concerns a damage-evolution setting, not
clinical performance.} Neither a smaller force nor a smaller displacement
alone establishes an improvement.

\section{Discussion}
\label{sec:discussion}

\subsection{Interpreting Tissue Effects}

The intended cut and the response of the remaining tissue provide the
context for interpreting the outputs. Tool loads describe the overall
interaction, strain describes local deformation, and \rev{the fixed-cohort statistic summarizes displacement
relative to the initial configuration. Its inclusion of last-saved
inactive-particle positions must be retained in the interpretation.} Figures~\ref{fig:reference} and~\ref{fig:comparison} organize these
quantities by common configuration or event. Their value is in the
correspondence between them, not in treating any one quantity as a
universal quality score.

\rev{The matched rate comparison tests post-initiation evolution, not
clinical superiority or a shift in the damage-onset criterion.}
A smaller force or displacement can accompany \reviewchange{less cut formation} and is
not, by itself, an improved tissue outcome. Relating mechanical response
to injury requires independent evidence, as in studies comparing
controlled loading with tissue histology~\cite{khan2021stress}. The
present outputs support conditional descriptions of tissue effects,
not an absolute safe/unsafe classification.

\subsection{Scope}

The isotropic material and scalar damage history are a compact
representation. Edge-localized initiation does not describe spontaneous
remote tearing; zero damage elsewhere is therefore not proof of tissue
preservation. \reviewchange{Delayed erosion removes active mass and energy; without
the optional barrier, shared grid support may retain coupling across
a narrow opening. The single-field barrier can lose affine
completeness for unfavorable crack/grid alignment, so its availability
does not establish general discontinuity accuracy.}
\rev{The rate limit is a model setting here, not an experimentally
calibrated tissue-fracture kinetic parameter. The matched rate runs do
not separately identify the effects of stress degradation and removal.
The archived batch recorded finite contact/G2P warnings; completion and
consistent code provenance do not establish error-free contact. The two
particle samplings support a repeated directional trend, not a
convergence claim.} The present study is numerical; tissue-specific
physical validation remains necessary.

\section{Conclusion}
\label{sec:conclusion}

STC-MPM links soft-tissue deformation, progressive failure, and
\reviewchange{configurable post-failure treatment with continued tool interaction.}
Material history governs
failure, and the resulting loss of load-carrying capacity affects
subsequent loading and movement. The framework provides tool reactions,
local tissue response, and \rev{material-region displacement} from the same
evolving body without a pre-inserted cutting interface. \rev{In matched
scalpel runs, a lower damage-rate limit left the first recorded damage
time unchanged but delayed peak resistance and first deactivation,
while increasing local displacement at maximum insertion. The results
illustrate how post-initiation failure evolution changes the subsequent
mechanical response under the same tool trajectory.} Its focus is a
coupled model for studying both cut formation and the accompanying
tissue effects, rather than prescribing a geometric incision or a
tissue trajectory independently.

\FloatBarrier

\end{document}